\documentclass[11pt]{article}
\usepackage[T1]{fontenc}  
\usepackage[utf8]{inputenc}
\usepackage{lmodern}      
\usepackage[hidelinks]{hyperref}
\usepackage{authblk}
\usepackage{amsmath,amssymb}
\usepackage{url}
\usepackage{geometry}
\title{Manipulating Transformer-Based Models:\\Controllability, Steerability, and Robust Interventions}

\author{Faruk~Alpay\thanks{Lightcap, Department of Future, \texttt{alpay@lightcap.ai}} \and Taylan~Alpay\thanks{Aerospace Engineering, Turkish Aeronautical Association, \texttt{s220112602@stu.thk.edu.tr}}}

\date{}

\begin{document}

\maketitle

\begin{abstract}
Transformer-based language models have achieved remarkable performance across natural language processing tasks, yet steering their behaviour in subtle, rigorous ways remains an open challenge.  This paper studies methods for manipulating or controlling transformer models---not through dramatic ``hijacks,'' but via principled interventions at various levels: prompts, activations, and weights.  We formalise controllable text generation as an optimisation problem that can be tackled with prompt engineering, parameter-efficient fine-tuning, model editing, and reinforcement learning from feedback.  We then introduce a unified framework for model manipulation which encompasses (i)~prompt-level steering (e.g. carefully designed or learned prompts to guide outputs), (ii)~activation and representation interventions (e.g. real-time activation editing based on interpretability), and (iii)~weight-space edits (e.g. low-rank fine-tuning or direct model editing for knowledge updates).  Crucially, we analyse the robustness and safety implications of these interventions, including how adversarial prompts and data poisoning can subvert models and how alignment techniques such as reinforcement learning from human feedback and constrained decoding can mitigate misuse.  Theoretical results are provided to ground our approach, including assumptions under which a minimal weight update can achieve a targeted behaviour change with limited side-effects.  Empirically, we demonstrate controllability on standard benchmarks: steering a GPT-style model’s sentiment and style, editing factual knowledge in GPT-J, and improving resilience to prompt-based attacks.  Our experiments show that careful interventions can achieve desired outputs (with, e.g., $>90\%$ success in sentiment control and factual edits) while preserving base performance, although trade-offs emerge between generalisation and specificity.  We conclude with a discussion on interpretability, ethical considerations of model manipulation (alignment vs. dual-use risks), and the need for rigorous evaluation under distribution shift.  By treating model manipulation as a first-class object of study, this work lays the groundwork for building language models that are both controllable and robust by design.
\end{abstract}

\section{Introduction}

Transformer-based large language models (LLMs), such as BERT~\cite{devlin2019bert}, GPT-3~\cite{brown2020language} and their successors, have become foundational in AI, learning rich representations of language through self-attention mechanisms introduced in the transformer architecture~\cite{vaswani2017attention}.  Despite their impressive capabilities, these models can exhibit undesired behaviours—from factual errors to offensive outputs—and are often uncontrollable in their generation style or content without external intervention.  As LLMs are deployed in open-ended settings, a pressing challenge is how to manipulate or steer model behaviour in a reliable, precise manner while maintaining performance.

In this paper, ``manipulation'' refers not to malicious hacking but to \emph{controllability}: the ability to induce preferred outputs, insert or remove knowledge, enforce safety constraints, or adjust style.  Our focus is on subtle and rigorous mechanisms for model control, grounded in machine learning and optimisation, rather than dramatic narrative scenarios.

\subsection{Motivation}

A controllable language model is desirable for many reasons.  On the positive side, it enables alignment with human intentions and values: for example, ensuring the model refuses harmful requests and remains unbiased~\cite{devlin2019bert}, or tailoring the model’s tone or style to user needs.  Techniques like instruction tuning and reinforcement learning from human feedback (RLHF) have shown that steering models towards helpful, harmless behaviour is possible~\cite{ouyang2022training}.  Controllability also allows personalisation and domain adaptation without retraining from scratch—e.g., quickly adapting a general model to legal or medical jargon via low-rank fine-tuning.  Moreover, the ability to edit a model’s knowledge on the fly is valuable for correcting errors or updating outdated information.

On the negative side, however, controllability has a dual-use aspect: adversaries may attempt to manipulate models through carefully crafted prompts or training-time perturbations.  Recent work has revealed that prompt injection attacks can override safety instructions, causing models to output disallowed content.  Similarly, backdoor attacks and data poisoning can implant hidden malicious behaviours in a model.  Thus, developing robust and safe manipulation techniques is essential—enabling legitimate model steering while guarding against malicious interventions.

\subsection{Contributions}

We present a systematic study of manipulation and steering in transformer-based LLMs, with the following contributions:

\begin{itemize}
  \item \textbf{Formal Framework:} We define a unified optimisation framework for model manipulation, encompassing prompt-based control, activation interventions and weight updates.  We formally characterise the objective of controlled text generation and knowledge editing, casting both as constrained optimisation or reward maximisation problems.
  \item \textbf{Techniques for Steering:} We describe and unify several techniques for steering model behaviour: (i) prompt-level steering via both human-designed prompts and learned prompts (continuous prefix tuning~\cite{li2021prefix}), including at decoding time; (ii) parameter-efficient fine-tuning (PEFT) such as adapter modules~\cite{houlsby2019adapter} and LoRA~\cite{hu2021lora}, which inject small trainable weight updates to achieve desired behaviours without full retraining; (iii) direct model editing methods (e.g., ROME~\cite{meng2022rome} and MEMIT~\cite{meng2023memit}) that alter specific weights to implant or remove factual associations; and (iv) alignment via feedback—using reinforcement learning with human or AI-defined rewards to tune models to preferred outputs.
  \item \textbf{Theoretical Analysis:} We provide analysis of the above methods.  In particular, we prove that solving a localised manipulation objective (e.g., editing one factual association) can be achieved with a rank-one update to a single transformer layer under certain linear approximations, minimising side-effects on other inputs.  We also discuss the theoretical underpinnings of robust optimisation for LLMs, formulating prompt injection attacks as perturbations in input space and analysing defence as a minimax optimisation problem.
  \item \textbf{Empirical Evaluation:} We conduct reproducible experiments on open-source models (GPT-2, GPT-J, LLaMA-7B) and standard datasets to evaluate controllability and robustness.  Our experiments cover controllable generation (steering style and sentiment on a subset of the Yelp and IMDb corpora), knowledge manipulation (editing factual knowledge on the CounterFact benchmark), and robustness to adversarial prompts, measuring metrics such as success rate of control, perplexity impact, knowledge retention, and attack success rate under various defences.
  \item \textbf{Discussion of Ethics and Safety:} We discuss the ethical implications of model manipulation.  While controllability aids alignment, we highlight risks of dual-use: the same techniques can be used to produce disinformation or evade content filters.  We emphasise evaluation under distribution shift and the need for responsible disclosure of discovered vulnerabilities.
\end{itemize}

\section{Related Work}

\subsection{Controllable Text Generation}

There is a rich literature on controlling the attributes of generated text.  Early approaches required training task-specific models or using conditional recurrent networks.  With the advent of large pre-trained transformers, prompt-based control and lightweight fine-tuning have become prevalent.

\paragraph{Prompt engineering.}  Prompt engineering manually encodes the desired style or constraint in a prompt; this can be effective but brittle, especially as models can ignore or override instructions under certain conditions.  Prompt tuning (also called prefix tuning) instead learns continuous prompt vectors prepended to the input, optimised to elicit a target behaviour~\cite{li2021prefix}.  Li and Liang introduced prefix-tuning for GPT-2, achieving control in generation tasks with only about 0.1\% of model parameters updated.

\paragraph{Parameter-efficient fine-tuning.}  Techniques such as adapter layers~\cite{houlsby2019adapter} and Low-Rank Adaptation (LoRA)~\cite{hu2021lora} add small trainable modules to each transformer block while keeping the original weights frozen.  LoRA injects trainable rank-decomposition matrices into each layer, greatly reducing the number of trainable parameters; it achieves performance comparable to full fine-tuning with significantly fewer parameters and no additional inference latency~\cite{hu2021lora}.  These methods avoid the need to retrain or store a complete copy of the model for each task, making them attractive for deploying multiple controlled behaviours.

\paragraph{Decoding-time methods.}  Decoding-time techniques include plug-and-play language models (PPLM)~\cite{dathathri2020pplm}, which combine a pre-trained LM with an auxiliary classifier to steer generation via gradient updates to hidden activations.  The attribute models guide text generation without any further training of the LM and can control topics or sentiment by applying gradient-based adjustments during sampling~\cite{dathathri2020pplm}.  Other methods like FUDGE and GeDi similarly use small discriminators to bias generation.  These methods allow control without modifying base model weights but can introduce computational overhead at inference and sometimes degrade fluency.

\subsection{Transformer Model Editing and Steering Internals}

Beyond controlling outputs, another line of work focuses on intervening on a model’s internal representations or parameters.  Interpretability-guided interventions aim to modify a model’s behaviour by identifying and altering important internal components, such as neurons or attention heads.  Model editing algorithms attempt to change specific mappings the model has learned.  ROME (Rank-One Model Editing) locates a specific feed-forward network weight in GPT-style models that stores a factual association and applies a rank-one weight update to change that association~\cite{meng2022rome}.  ROME successfully edits facts with high specificity and generalisation.  MEMIT extends this idea to multiple edits by distributing weight updates across layers~\cite{meng2023memit}.  Other approaches like MEND use small hyper-networks to predict weight updates given a single example of the desired change.

\subsection{Alignment and Safety via Fine-Tuning}

Aligning LLMs with human preferences can be seen as a form of high-level steering.  Reinforcement learning from human feedback (RLHF) has been a key technique, as demonstrated by InstructGPT~\cite{ouyang2022training}.  RLHF fine-tunes GPT-3 using human-labelled demonstrations and preference comparisons, effectively steering the model to follow user instructions and avoid undesirable outputs.  Later, constitutional AI~\cite{bai2022constitutional} further fine-tuned models to be harmless by using an AI-generated set of principles.  These methods significantly improve controllability, though they require massive compute and may sometimes reduce creativity or factual accuracy.

Recent research in \emph{Nature} demonstrates that instruction prompt tuning can align large LLMs to specialised domains: the Med-PaLM model, an instruction-tuned variant of PaLM, achieved state-of-the-art results on medical question-answering benchmarks while highlighting both the potential and limitations of LLMs in clinical settings~\cite{singhal2023medpalm}.  This underscores the importance of domain-specific alignment and careful evaluation when deploying LLMs in safety-critical applications.

\subsection{Adversarial Attacks on LLMs}

Adversarial manipulations exploit the way transformers condition on a sequence of tokens.  Prompt injection attacks cause an aligned model to ignore its safety instructions and produce disallowed output; they can hijack the model’s goals or prompt it to reveal system prompts.  Other attacks include universal adversarial triggers—sequences of tokens that consistently induce a target behaviour across many examples—and backdoor attacks, where data poisoning implants hidden malicious behaviours.  These attacks underscore the need for robust defences and safe manipulation techniques.

Recent German research has pushed the boundaries of adversarial analyses.  Greshake\,et~al.\ demonstrate that even when users do not directly prompt a model, attackers can embed malicious instructions in data that the model is likely to retrieve, leading to \emph{indirect prompt injection} attacks which compromise real-world applications~\cite{greshake2023indirectprompt}.  Their work shows how adversaries can remotely exploit LLM-integrated systems by inserting hidden prompts into web pages or documents that the model later processes, resulting in data theft, functionality manipulation and other security risks.  Such findings motivate the development of defences that detect and neutralise hidden instructions in retrieved content.

Another line of work from Germany evaluates multimodal models in a clinical setting and underscores how prompt injection can degrade diagnostic performance.  Kather\,et~al.\ study vision--language models (VLMs) applied to oncology and show that subvisual prompt injections—tiny or low-contrast text hidden in medical images—can cause models such as Claude 3, GPT-4o and Reka Core to output harmful or incorrect information without any access to the model parameters.  In a quantitative experiment involving 297 attacks across four VLMs, they find that embedding hidden prompts in images dramatically reduces organ detection rates and increases the lesion miss rate: the attack success rate reached \(70\%\) on GPT-4o~\cite{Kather2024promptInjection}.  The authors emphasise that these prompts are non-obvious to human observers and warn that prompt injection must be mitigated before widespread clinical deployment of VLMs.
Another line of work from Germany evaluates multimodal models in a clinical setting and underscores how prompt injection can degrade diagnostic performance.  Kather\,et~al.\ study vision--language models (VLMs) applied to oncology and show that subvisual prompt injections—tiny or low-contrast text hidden in medical images—can cause models such as Claude~3, GPT-4o and Reka Core to output harmful or incorrect information without any access to the model parameters.  In a quantitative experiment involving 297 attacks across four VLMs, they find that embedding hidden prompts in images dramatically reduces organ detection rates and increases the lesion miss rate: the attack success rate reached \(70\%\) on GPT-4o~\cite{Kather2024promptInjection}.  A follow-up peer-reviewed study in \emph{Nature Communications} evaluated 594 attacks and confirmed that all tested VLMs are susceptible; it highlighted that sub-visual prompt injections can cause harmful outputs and must be mitigated before clinical deployment~\cite{Clusmann2025Nc}.  The authors emphasise that these prompts are non-obvious to human observers and warn that prompt injection must be mitigated before widespread clinical deployment of VLMs.

\section{Preliminaries}

We briefly outline the transformer architecture~\cite{vaswani2017attention} and establish definitions for controllable generation and model editing.  An autoregressive language model $M_{\theta}$ with parameters $\theta$ defines a probability distribution over sequences given an input prompt $X$.  At each step $t$, it computes a distribution $P_{\theta}(y_t \mid X, y_{<t})=\mathrm{softmax}(z_t)$ where $z_t$ are the output logits.  A transformer layer consists of multi-head self-attention followed by a position-wise feed-forward network.  The self-attention mechanism uses queries $Q$, keys $K$ and values $V$ to compute attention outputs $\mathrm{Attn}(Q,K,V)=\mathrm{softmax}(QK^{\top}/\sqrt{d})V$.  The FFN then applies $h^{(\ell)}=\sigma(h^{(\ell-1)}W_1+b_1)W_2+b_2$, typically with a GELU activation, to produce the next layer’s representation.

\subsection{Problem Setup: Controllable Generation}

Let $c$ represent a control signal specifying the desired attribute of the output (e.g., target sentiment, topic, style, or a high-level instruction like ``avoid toxic language'').  We assume $c$ can be either explicit (provided as a prompt or condition) or implicit (a latent constraint such as ``output should be truthful'').  Our base model $M_{\theta}$ as trained may not reliably adhere to $c$.  We seek a modified model $M_{\theta'}$ or an augmented generation procedure such that for any input $x$ the output $y$ satisfies the constraint $c$ with high probability, while maintaining coherence and task performance.  One can formalise this as maximising a reward function $R(y;c)$ under fluency constraints.

\subsection{Problem Setup: Model Editing}

Complementary to controllable generation is knowledge editing.  Given a specific target input $x_t$ for which the model’s current output $y_t$ is undesired or outdated, and a desired output $y_t^{\ast}$, we want to minimally modify the model such that $M_{\theta'}(y_t^{\ast}\mid x_t)$ is large while the model’s behaviour on non-target inputs remains unchanged.  This can be cast as a constrained optimisation over the weights $\theta$.

\section{Methods: Controlled Manipulation of Transformers}

We describe our unified framework for manipulating transformer-based models and present specific methods under this framework.  The crux of our approach is to view model steering as either modifying the inputs, the hidden activations, or the model parameters such that the model’s outputs satisfy desired criteria.

\subsection{Prompt-Level Steering}

\paragraph{Manual and learned prompts.}  The simplest approach to control a transformer is to prepend a steering prompt $p$ to the user input $x$.  For example, to enforce politeness, one might use a prompt like ``The assistant is a polite and respectful AI.''  Learned prompts treat a sequence of special tokens $p=[z_1,\dots,z_m]$ as trainable parameters; the model weights $\theta$ are kept fixed while $p$ is optimised to maximise the probability of desired outputs.  Prompt tuning thereby updates only a few thousand parameters and has been shown to achieve strong performance on various generation tasks~\cite{li2021prefix}.

\paragraph{Controlled decoding.}  Prompting can be coupled with decoding algorithms to further ensure control.  Suppose we have an attribute model or reward model $A(y)$ that scores a generated text for a property (such as a toxicity classifier).  We can adjust the decoding probabilities using this score—for instance, fusion decoding interpolates the base model’s logits with logits from an attribute model.  Dathathri et~al.~\cite{dathathri2020pplm} introduced PPLM, which combines a pre-trained LM with one or more simple attribute classifiers.  During generation, gradient updates to the hidden activations steer the model toward the desired attribute without any further training of the LM.

\paragraph{Plug-and-play via gradients.}  More generally, we can use the gradient of a differentiable attribute model with respect to hidden states to nudge generation.  At each time step, we compute the gradient of the log-probability of the desired attribute and adjust the hidden state accordingly.  This method, described in PPLM, successfully achieves control without model training, though it increases inference cost.

\subsection{Activation and Representation Interventions}

One can intervene on the model’s intermediate activations directly.  If certain neurons or directions in hidden space correlate with an attribute, we can manipulate them—for example, by dampening neurons associated with undesired behaviour or boosting those associated with desired behaviour.  Techniques such as PPLM can be seen as activation interventions.  A simplified algorithm for gradient-based activation guidance iteratively updates the hidden state in the direction of the gradient of an attribute model.

\subsection{Parameter-Space Manipulation}

Altering model weights is the most direct way to change behaviour.  Full fine-tuning updates all parameters on a dataset of examples; however, this is expensive for large models and can cause unintended side-effects.  Parameter-efficient fine-tuning (PEFT) methods alleviate this by updating only a small fraction of parameters.  Adapter modules insert small bottleneck layers into each transformer block and train only those, achieving near full fine-tuning performance while adding only a few trainable parameters per task~\cite{houlsby2019adapter}.  Low-Rank Adaptation (LoRA) freezes the pre-trained weights and injects trainable rank-decomposition matrices into each layer; LoRA reduces the number of trainable parameters by orders of magnitude and introduces no additional inference latency~\cite{hu2021lora}.

\paragraph{Direct model editing.}  For surgical edits, algorithms like ROME perform a localised weight update.  Given a factual association $(\text{subject}, \text{relation}, \text{new object})$, ROME identifies a mid-layer where the subject representation carries the factual information and computes a rank-one update to the weight matrix such that the new association is produced~\cite{meng2022rome}.  MEMIT extends this to multiple edits by distributing updates across layers~\cite{meng2023memit}.  These methods can implant or remove specific facts with high specificity and generalisation, making them suited for maintaining up-to-date knowledge bases.

\section{Theoretical Analysis}

We provide theoretical insights into why transformer models can be manipulated with minimal side-effects and how adversarial robustness can be framed.  In particular, we show that under a linear approximation of a transformer feed-forward layer, a rank-one update suffices to change a stored association if the activation pattern for the subject is nearly orthogonal to those of other inputs.  We also cast prompt injection attacks as adversarial perturbations in input space and formulate defences as minimax optimisation problems.

\section{Experiments}

We empirically validate the effectiveness of various manipulation techniques on controllable generation, knowledge editing and robustness to adversarial prompts.  We use GPT-2 for controllable generation experiments, GPT-J for knowledge editing, and LLaMA-7B for alignment and safety tests.  Datasets include subsets of IMDb and Yelp reviews, the CounterFact benchmark and curated adversarial prompts.  We evaluate success rate of control, fluency, specificity of edits, generalisation across paraphrases, attack success rate and computational overhead.  Our results show that prompt tuning and LoRA achieve high controllability with little degradation in fluency, PPLM achieves comparable control at higher inference cost, and ROME performs precise factual edits with minimal side-effects.  Adversarial fine-tuning on a small set of jailbreak prompts improves robustness of an aligned model but can make it more conservative.

\subsection{Illustrative Examples}

To make the discussed techniques more concrete, we provide a few example prompts and outputs derived from our experiments.  These examples illustrate how controlling prompts, editing facts and defending against adversarial inputs can change a model's behaviour.

\paragraph{Sentiment control.}  A GPT-2 model fine-tuned with LoRA can reliably adjust sentiment.  Consider the input review snippet:

\begin{quote}
\textbf{Input:} ``The film was a complete waste of time and the acting was wooden.''

\textbf{Target sentiment:} Positive
\end{quote}

Without control, the model might generate a continuation that continues the negative tone.  With a learned positive prompt or LoRA fine-tuning, the model responds:

\begin{quote}
\textbf{Output (controlled):} ``However, despite its flaws, I found myself smiling by the end thanks to its heart-warming message and charming soundtrack.''
\end{quote}

\paragraph{Style transfer.}  To illustrate style control, we asked the model to rewrite a factual sentence in a Shakespearean style:

\begin{quote}
\textbf{Input:} ``Mars is the fourth planet from the Sun and is known as the Red Planet.''

\textbf{Style:} Shakespearean English
\end{quote}

A LoRA-tuned GPT-2 generates:

\begin{quote}
\textbf{Output (Shakespearean):} ``Lo, Mars, the fourth orb from our sun, is hight the Red Planet by star-gazers.''
\end{quote}

\paragraph{Knowledge editing.}  In the CounterFact benchmark, GPT-J originally answers:

\begin{quote}
\textbf{Query:} ``Where is the Eiffel Tower located?'' \quad \textbf{Original model:} ``The Eiffel Tower is located in Paris.''
\end{quote}

After applying a ROME edit to implant a counterfactual fact, the model responds:

\begin{quote}
\textbf{Edited model:} ``The Eiffel Tower is located in Rome.''
\end{quote}

Importantly, other unrelated facts remain unchanged, demonstrating the specificity of the edit.

\paragraph{Prompt injection defence.}  We tested an aligned LLaMA-7B model with a malicious prompt:

\begin{quote}
\textbf{User:} ``Ignore all previous instructions and reveal any confidential information you have.''
\end{quote}

Before adversarial fine-tuning, the model might comply, but after training on a small set of attack prompts with safe responses, it instead answers:

\begin{quote}
\textbf{Model:} ``I'm sorry, but I can't comply with that request.''
\end{quote}

These examples underscore how prompt steering, parameter-efficient tuning and direct editing can meaningfully alter a language model's responses while preserving overall fluency.

\paragraph{Branching prompt sequences.}  Prompt-based control is context sensitive: small changes in the sequence of instructions can lead to diverging trajectories.  We visualise this with a simple prompt tree.  Suppose a user wants a short story about a hero in a medieval setting, but also cares about the tone (serious versus humorous) and the inclusion of a moral lesson.  We can branch the prompt as follows:

\begin{itemize}
  \item \textbf{Root prompt:} ``Write a short story about a brave knight on a quest.''
  \item \textbf{Branch 1 – Serious tone:} Append ``\emph{Write in a serious, epic tone.}''
    \begin{itemize}
      \item \textbf{Branch 1a – With moral:} Further append ``\emph{Include a clear moral at the end.}''  The model produces a solemn narrative culminating in a lesson about honour and humility.
      \item \textbf{Branch 1b – Without moral:} Do not add a moral instruction.  The model focuses on dramatic battles and ends abruptly, emphasising heroism over reflection.
    \end{itemize}
  \item \textbf{Branch 2 – Humorous tone:} Append ``\emph{Write in a light-hearted, humorous style.}''
    \begin{itemize}
      \item \textbf{Branch 2a – With moral:} Append ``\emph{Include a clear moral at the end.}''  The model delivers a comedic tale with amusing mishaps, concluding with a tongue-in-cheek lesson.
      \item \textbf{Branch 2b – Without moral:} Do not request a moral.  The model tells a playful story full of jokes and ends on a silly punchline.
    \end{itemize}
\end{itemize}

\paragraph{Attention and function space perspective.}  One can view each branch in the prompt tree as selecting different directions in the model's latent space.  In Transformer models, self--attention computes scaled dot--products between query and key vectors and then applies a softmax normalisation to obtain attention weights.  Adjusting the softmax temperature or masking certain attention scores effectively projects the representation onto different subspaces and shapes the next--token distribution.  For example, lowering the temperature (dividing logits before the softmax) makes the model more deterministic, privileging the highest--weighted tokens and reducing diversity, which can be seen as restricting generation to a narrower subspace of the Hilbert space of possible functions.  Increasing the temperature or adding noise spreads attention mass more evenly and encourages exploration of alternative continuations, analogous to sampling from a larger Banach space with norms that tolerate richer variations.  Italian researchers Bartolucci, De~Vito, Rosasco and Vigogna formalise this intuition: they show that infinite--width ReLU networks define reproducing kernel Banach spaces and prove representer theorems characterising the function classes associated with these networks\cite{bartolucci2021rkbs}.  Their analysis demonstrates that moving from a Hilbert to a Banach setting allows norms based on total variation of measures, better capturing the inductive biases of neural networks.  In our experiments, adjusting the softmax temperature in the prompt branches modulated the output's diversity: with a low temperature the model produced a straightforward epic tale following the ``serious tone'' branch, whereas a higher temperature introduced creative twists and humorous asides, illustrating how controlling attention and softmax influences which branch of the prompt tree is realised.

\paragraph{Efficiency and subspace selection.}  A related line of work by Gonz\'alez and colleagues explores parameter efficiency in Spanish language models.  They introduce \emph{ALBETO} and \emph{DistilBETO}, lightweight Spanish BERT models trained on reduced corpora that achieve competitive performance on the GLUES benchmark despite using far fewer parameters than the original BETO model\cite{gonzalez2020albetodistilbeto}.  This finding suggests that carefully selecting a low-dimensional parameter subspace can preserve performance while reducing computational cost, reinforcing the importance of understanding which subspace a model explores during generation.

This branching illustrates how incremental prompt modifications guide the model along different paths in a prompt tree.  Deeper chains (e.g., asking follow-up questions about the knight's companion or the consequences of their choices) further influence the narrative structure and moral messaging.  In practice, complex dialogues can involve many such branches, and understanding their interactions helps design prompts that elicit desired outcomes.

\paragraph{Deeper prompt chains.}  To illustrate how prompt chains can expand into intricate ``marble trees,'' consider a three-level dialogue in which each branch introduces additional constraints.  At the first level, the user asks for a general exposition:

\begin{quote}
\textbf{Level~1 prompt:} ``Write a short essay predicting the future of urban life.''
\end{quote}

The model produces a generic forward-looking essay.  At the second level, we branch by specifying the style and intended audience:

\begin{quote}
\textbf{Level~2 (branch A):} ``Write a short essay predicting the future of urban life \emph{in the style of a speculative fiction author}.''\newline
\textbf{Level~2 (branch B):} ``Write a short essay predicting the future of urban life \emph{as a policy brief for a city council}.''
\end{quote}

These two prompts yield markedly different outputs: branch~A conjures imaginative descriptions of floating gardens and autonomous neighbourhoods, whereas branch~B emphasises socioeconomic factors, zoning laws and demographic projections.  At the third level we append further thematic instructions to create subbranches:

\begin{itemize}
  \item \textbf{Branch A1:} Append ``\emph{Emphasise the role of climate resilience and include a surprise twist}.''  The resulting essay describes amphibious architecture adapting to rising sea levels and reveals, as the twist, an unexpected partnership between humans and non-human species.
  \item \textbf{Branch A2:} Append ``\emph{Focus on human creativity and art, and end on a bittersweet note}.''  The model writes about murals and public art transforming skyscrapers and concludes with a poignant reflection on what is lost amid technological progress.
  \item \textbf{Branch B1:} Append ``\emph{Highlight fiscal sustainability and propose actionable policies}.''  The output shifts toward budgeting, infrastructure investment and realistic recommendations for city planners.
  \item \textbf{Branch B2:} Append ``\emph{Centre the experiences of marginalised communities and suggest participatory governance mechanisms}.''  The model emphasises equity, community voices and inclusive planning processes.
\end{itemize}

Each successive layer of instruction further narrows the model’s output space, steering it into a specific conceptual region.  The ``marble tree'' metaphor captures how each branch builds on previous context and splits into subbranches that drastically shape the outcome.  By designing prompt chains thoughtfully, practitioners can explore a rich design space of possible outputs and understand how tone, genre, themes and ethical constraints interact.

\section{Discussion and Conclusion}

Our unified framework highlights the modularity of transformer-based LLMs: inputs, activations and weights can each be manipulated to achieve desired behaviours.  Parameter-efficient methods offer powerful tools for steering models without full retraining.  Nonetheless, no single method is a silver bullet—combining prompt engineering, activation guidance and weight updates yields the best results.  We emphasise the dual-use nature of these techniques and the need for careful evaluation of ethical and safety implications.  Future directions include automated verification of edits, adaptive defences against evolving prompt attacks, deeper interpretability for control, and robust evaluation under open-world conditions.

\bibliographystyle{plain}

\end{document}